\def\year{2022}\relax
\documentclass[letterpaper]{article} 
\usepackage{aaai22}  
\usepackage{times}  
\usepackage{helvet}  
\usepackage{courier}  
\usepackage[hyphens]{url}  
\usepackage{graphicx} 
\usepackage{natbib}  
\usepackage{caption} 
\DeclareCaptionStyle{ruled}{labelfont=normalfont,labelsep=colon,strut=off} 
\usepackage{algorithm}
\usepackage{algorithmic}
\usepackage{algorithm}
\usepackage{algorithmic}
\usepackage{amssymb}
\usepackage{amsfonts}
\usepackage{amsthm}
\usepackage{amsmath}
\usepackage{multirow}
\usepackage{makecell}

\usepackage[colorlinks,
            linkcolor=blue,
            anchorcolor=blue,
            citecolor=blue]{hyperref}

\usepackage{newfloat}
\usepackage{listings}
\newtheorem{definition}{Definition}
\newtheorem{example}{Example}

\usepackage{newfloat}
\usepackage{listings}
\floatstyle{ruled}
\newfloat{listing}{tb}{lst}{}
\floatname{listing}{Listing}
\title{A Concept and Argumentation based Interpretable Model in High Risk Domains}
\author{
    Haixiao Chi\textsuperscript{\rm 1}, Dawei Wang\textsuperscript{\rm 2}, Gaojie Cui\textsuperscript{\rm 2}, Feng Mao\textsuperscript{\rm 2}, Beishui Liao\textsuperscript{\rm 1}\thanks{The corresponding
author of this paper. Contact him at baiseliao@zju.edu.cn.}
}
\affiliations{
    \textsuperscript{\rm 1} Zhejiang University, Hangzhou, China\\
    \textsuperscript{\rm 2} Alibaba Group, Hangzhou, China\\
}

\usepackage{bibentry}

\begin{document}

\maketitle

\begin{abstract}

Interpretability has become an essential topic for artificial intelligence in some high-risk domains such as healthcare, bank and security. For commonly-used tabular data, traditional methods trained end-to-end machine learning models with numerical and categorical data only, and did not leverage human understandable knowledge such as data descriptions. Yet mining human-level knowledge from tabular data and using it for prediction remain a challenge. Therefore, we propose a concept and argumentation based model (CAM) that includes the following two components: a novel concept mining method to obtain human understandable concepts and their relations from both descriptions of features and the underlying data, and a quantitative argumentation-based method to do knowledge representation and reasoning. As a result of it, CAM provides decisions that are based on human-level knowledge and the reasoning process is intrinsically interpretable. Finally, to visualize the purposed interpretable model, we provide a dialogical explanation that contain dominated reasoning path within CAM. Experimental results on both open source benchmark dataset and real-word business dataset show that (1) CAM is transparent and interpretable, and the knowledge inside the CAM is coherent with human understanding; (2) Our interpretable approach can reach competitive results comparing with other state-of-art models.


\end{abstract}
\section{Introduction}
For decision-making tasks in high-risk domains, machine learning (ML) methods are required to have high level of interpretability. 
Many feature importance based post-hoc explainable methods, such as SHapley Additive exPlanations (SHAP) \cite{shap} and Local Interpretable Model-Agnostic Explanations (LIME) \cite{lime}, are proposed to explain black box models. However, \cite{fool} showed that feature importance based explanation can neither reflect the real behavior of the black model nor improve human understanding of the model. Thus, interpretable models have been an increasingly active research direction, and high order Generalized Additive Models (GAMs) such as Explainable Boosting Machine (EBM) \cite{ebm} and NODE-GAM \cite{nodegam} are purposed to provide analysis on individual features or interaction between two features toward the prediction target. Additionally, a line of researches have focused on the high-order features and feature grouping methods \cite{
autocross, deep}. However, the aforementioned methods are purely data-driven and did not include domain knowledge from expertise, and sometimes the resulted feature interactions from those methods are hard to be understood by humans.

To improve the interpretability of the above methods, argumentation-based methods are purposed to increase human understanding and trust of the model by injecting the human-level knowledge during decision-making stage. Because formal argumentation, as a formalism for representing and reasoning with knowledge \cite{handbookargumentation}, are capable of providing various ways for justifying and explaining why a claim or a decision is made \cite{Contrastive,explanation}. Among them, quantitative argumentation frameworks (QAF) has a greater advantage over qualitative argumentation when combined with data-driven methods. To construct the argumentative structure, majorities of the existing augmentation-based methods are bespoke to a specific problem \cite{ant, tweet} and depend on the knowledge from in-domain expertise, which significantly limits the usage of these methods since the specific argumentation structures are hard to migrate to a new problem.

\begin{figure}
\centering
\includegraphics[scale=0.33]{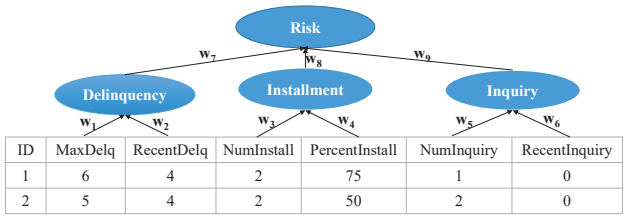}
\caption{An example of concept abstraction from tabular data for risk estimation}
\label{fig:example_t}
\end{figure}

One approach to address the above issue is to automate the argumentation structure generation process with data miming techniques using orthogonal in-domain knowledge information from external data. For tabular data in high risk domain, data descriptions are one of the great resources for in-domain knowledge since in-domain expertise needs to make decisions depends on raw features value directly \cite{DBLP:conf/aies/LakkarajuKCL19}. In this paper, we propose a concept and argumentation based model (CAM) that generates human-understandable concepts by mining data descriptions automatically and representing the generated concepts and the reasoning paths with argumentation structure. To illustrate CAM, Fig. \ref{fig:example_t} shows a concrete example from a real word high risk application \cite{fico} : to explain the decision-making process, concept \emph{Installment} is generated automatically from the two underlying features based on their similar data descriptions, and the processes are repeated twice to generate concepts \emph{Delinquency} and \emph{Inquiry}. On top of them, concept \emph{risk} is generated 
to represent the final decision-making process. The resulted concept-based knowledge can be properly represented and reasoned using QAF: each concept can be viewed as an abstract argument, while the 
inter-concept edges can be understood as supports or attacks between arguments, and a quantitative argumentation-based field-wild leaning algorithm is designed to evaluate and filter the generated concepts. 

With quantitative argumentation, CAM can be represented as stacked QAFs with weighted edges that represent the inter-concept relationship strength, and a quantitative argumentation-based method is designed as the reasoning machine inside the stacked QAFs to conduct knowledge reasoning by aggregating the strengths of lower-level concepts or features to the strengths of higher-level concepts. The reasoning machine can output a knowledge reasoning path that can be used as global model interpretation. As a result of it, CAM can be treated as an interpretable white-box model since the decision-making process is transparent to the users with the visualization of the reasoning path in form of dialogical tree.

\textbf{Our contributions.} To summarize, our contributions are listed as follows:
\begin{itemize}
\item Conceptually, we propose CAM to automatically generate and evaluate concepts from tabular data, and utilize quantitative argumentation to represent and reason on concepts. Furthermore, we provide explanation as the key reasoning path within CAM.
 
\item Empirically, we conduct CAM on both open source benchmarks and real-word bususiness datasets in high-risk domains. Experimental results show that (1) CAM is competitive compared to other ML models; (2) CAM is both global and local interpretable, and the knowledge inside the model is coherent with human understanding.




\end{itemize}

\section{Related work}

\paragraph{Explainable artifical intelligence works for tabular data} For tabular data, a group of post-hoc methods explain models by computing or approximating the feature importance, such as LIME and SHAP. But, the explanation can be complex and difficult to be understood by humans because the granularity of the explanation is too fine \cite{beyond}. Recently, researchers have focused on the interaction of features rather than individual features by constructing interpretable model. NODE-GAM and EBM are a class of interpretable model that can provide analysis on individual features or interaction between two features. These two methods apply neural network models and boosting tree models, respectively, to fit a functional relationship between a single feature or an interaction feature with the output. However, \cite{badgam} argue that different fitting models can have different or even contradictory interpretations for the same data. This is because the fitting model model relies solely on data-driven and leads to overfitting. DANETs\cite{deep} are able to abstract higher-level tabular features by nerual network. But the higher-level features only contribute to the classification performance. The semantic within them is not completely explicit, which may lead to confusing features.   

\paragraph{Interpretble Concepts Mining} Recent researches have focused on generating high-level human concepts from data. TCAV \cite{tcav} and VCEC \cite{vcec} produced estimation of how important a concept is for the prediction. ProtoPNet \cite{protype} is trained to learn visual prototype vector and calculate similarity for prediction. ACE \cite{ace} proposed a method to automatically extract visual concept from certain class’s images. But all the above-mentioned methods are designed specifically for image data. Our goal is to make similar effort for interpretable tabular ML.

\paragraph{Quantitative argumentation-based work in explainable artificial intelligence }  Quantitative argumentation frameworks (QAF) are a knowledge representation formalism that can be used to solve decision problems in a very intuitive way by weighing up pro and contra arguments \cite{qaf}. QAF are are based on Bipolar Argumentation Frameworks (BAFs) \cite{baf} by quantifying the semantics of arguments and the relations between them. Many reasoning methods are proposed for evaluating their semantics, including DF-QuAD algorithm \cite{dfqaf}, O-QuAD algorithm \cite{ant}, Multi-layer Perception (MLP)-based algorithm \cite{mlpqaf} and etc. 
These models have better performance in solving real life problems, such as fraud detection \cite{ant}, opinion polling \cite{polling}, and review aggregation \cite{review}. However, the argumentation model rely on a concrete knowledge structure. It can be the inherent structure of the data, such as the tree conversation structure in social media \cite{tweet}, or constructed manually by human experts \cite{ant}. To the best of our knowledge, existing works have not introduce a method that can automatically construct argumentative tree from tabular data.

\section{Concepts and argumentation based model}

\begin{figure*}
\centering
\includegraphics[scale=0.45]{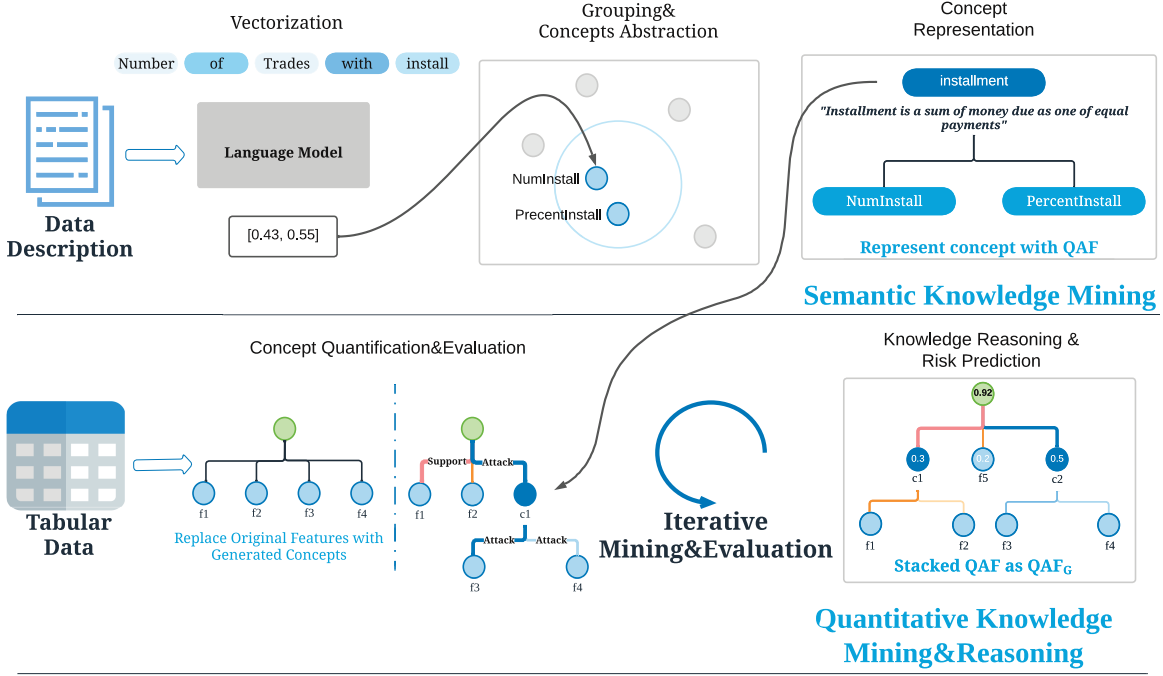}
\caption{The overall workflow of CAM. In CAM, we first mine semantic knowledge as concepts from data description and represent them in QAFs. Then, the quantitative knowledge is mined from underlying data for concept quantification and evaluation. Repeating these two step, we can built $QAF_G$ by stacking concepts in form of QAF for decision-making.}
\label{fig:cam}
\end{figure*}

\subsection{Knowledge definition and representation}

\subsubsection{knowledge in tabular data}


In tabular data, the knowledge may be divided into two categories: human-level knowledge and knowledge learned from data. The former can be data description which express the semantics of features in natural language and is intuitively understandable to human beings.
From a cognitive perspective \cite{concept}, we believe that a concept should be knowledge that abstracts common characteristics from a set of features or lower-level concepts, which should be consistent with human cognition, as shown in Fig. \ref{fig:example_t}.  
Another kind of knowledge can be learn from underlying data and expressed as the value of higher-level concepts relevant to the prediction targets which is aggregated by exploiting the correlative features or lower-level concepts in the groups. And a concept tree is constructed with concepts, their children and the correlation between them.

\subsubsection{Represent knowledge in quantitative argumentation framework} A QAF is a quadruple $\langle\mathcal{A}$, $E$, $\beta, \omega \rangle$ consisting of a set of arguments $\mathcal{A}$ , edges $E \in \mathcal{A} \times \mathcal{A}$ between these arguments, a function $\beta: \mathcal{A} \rightarrow [0,1]$ that assigns a $base\ score\ \beta(a)$ to each argument $a \in \mathcal{A}$ and a function $\omega: E \rightarrow \mathbb{R}$ that assigns a weights to each edge.  Furthermore, for every argument $a \in \mathcal{A}$, we let $Att(a) = \{(b,a) \in E\ |\ \omega(b,a)<0 \}$ and $Sup(a) = \{(b,a) \in E\ |\ \omega(b,a)>0 \}$

 
 A concept tree rooted by concept with the global semantics (denoted as $c_g$) can be represented in form of QAF as $QAF_{G}: \langle\mathcal{A}$, $E$, $\beta, \omega \rangle$, each argument $a$  in set $\mathcal{A}$ represents a concept $c$ or a feature $f$, edges $E \subseteq \mathcal{A} \times \mathcal{A}$ between these arguments describe the positive and negative correlation between concepts and features, a function $\beta: \mathcal{A} \rightarrow [0,1]$ that assigns a $base\ score\ \beta(a)$ to each argument $a \in \mathcal{A}$ which represents a concept, and a function $\omega: E \rightarrow \mathbb{R}$ that assigns a weights to each edge. ($\beta , \omega$ are not defined now until the field-wise learning algorithm is introduced.) It is worth noting that features as a unit of knowledge are also represented as arguments. However, in the QAF, the arguments representing features are at the leaf node positions, because the knowledge embedded in features has the finest granularity. In the tabular database, features have been mapped to the underlying data. Therefore the strength of the arguments representing the features can be obtained directly from the data without the need of the base score function to assign initial values. We denote features as $f \in \mathcal{F}$, and $\mathcal{F}$ is a set of features contained in tabular data.

\subsubsection{Definition of concept}
From the perspective of knowledge representation, the knowledge contained in a concept can be represented as two characteristics of the concept: a semantic-based and an argumentation-based characterisation. The former one can be described in form of natural language for human or semantic vector for machine, representing the knowledge learned from data description. The other can be represent in form of QAF rooted by this concept, in which the nodes $\mathcal{A}$ and edges $E$ in QAF are learned from data description and $\beta$ and $\omega$ are learned from underlying data. 

\begin{definition} \label{def-concept}
{\rm Let} $\mathcal{C}$ {\rm be a set of concepts, a concept} $c \in \mathcal{C}$.{\rm A} semantic-based characterisation of c {\rm is meaning} $m_c$ $\in \mathcal{M}$. {\rm An} argumentation-based characterisation of c {\rm is } $QAF_c \subseteq QAF_{G}$, {\rm where} $\mathcal{M}$ {\rm is a finite set of meanings of concepts, and }$QAF_c$ {\rm is the subtree of } $QAF_{G}$.
\end{definition}

A concept $c$ consists two characterisations $\langle m_c, QAF_c\rangle $, $m_c$ represents the semantic information and $QAF_c$ describes its argumentative information. As shown in Fig. \ref{fig:example_t}, we take concept 'Installment' as an example.

\begin{example}
{\rm concept} $c_{Installment}$ {\rm have the meaning} $m_{installment}$ {\rm as `installment is a sum of money due as one of several equal payments for something, spread over an agreed period of time'. And its argumentative structure} $QAF_{installment}=\langle\mathcal{A}$, $E$, $\beta, \omega \rangle${\rm , in which }$\mathcal{A}=\{c_{installment}, f_{NumInstall}, f_{PercentInstall}\}$, $E = \{ (f_{NumInstall}, c_{installment}), (f_{PercentInstall}, c_{installment}) \}$,  $\beta$ {\rm and }$\omega$ {\rm can be instantiated by learning from the underlying data.} 

\end{example}

\subsection{Knowledge acquisition: concept mining method}

In this section, we present the concept mining method which can be decoupled into two processes: (i) semantic knowledge mining, and (ii) quantitative knowledge mining, as shown in Fig. \ref{fig:cam}. Semantic knowledge mining is designed for automatically searching lower-level knowledge units (such as features and lower-level concepts) with similar meanings and abstract higher-level concepts from them. Then, quantitative knowledge mining is designed for mining the correlations between concepts and their children. 


\subsubsection{Semantic knowledge mining}
To simulate the process of abstracting concepts in human cognitive learning, we need to combine features with similar meaning and extract the same characteristics as the meaning of the generated concepts. To achieve this goal, three main procedures are necessary in Semantic knowledge mining: 1) Vectorization. 2) Grouping. 3) Abstraction.

\textbf{Vectorization} transfers the natural language information into vector space with the leverage of pretrained multi-lingual language model. In that way, the meaning of features or concepts is embedded into vector from natural sentences. And then \textbf{Grouping} process the semantic vectors into several groups by combine the features or concepts with high similarity. (In our task, every group contains two elements which can be features or concepts). Finally, in \textbf{Abstraction} process, the similar descriptive part of the natural statements in each group is extracted and vectorized as the meaning of newly generated higher-order concept. To be noticed, the structures of the new concepts are also mined in this process, which can be represented in form of  $\mathcal{A}$ and $E$ in their QAFs.

For example, in Fig.\ref{fig:example_t}, $m_{f_{NumInstall}}$ of $f_{NumInstall}$ in description of dataset is `number of trades with installment, installment is one of several equal payments for something, spread over an agreed period of time'. And $m_{f_{PercentInstall}}$ is described as `percents of trades with installment, installment is ...'. Through procedures of Vectorization and Grouping, $f_{NumInstall}$ and $f_{PercentInstall}$ are grouped together. Then, $c_{installment}$ is abstracted from them, and the meaning of $c_{installment}$ is represented as `installment is one of several equal payments for something, spread over an agreed period of time'. To make machine understand, $m_{c_{installment}}$ will also be the semantic vector. Also, $\mathcal{A}$ and $E$ of $QAF_{c_{installment}}$ is captured as showed in Example 1. 


\subsubsection{Quantitative knowledge mining}
After the process of semantic knowledge mining, as shown in Fig.\ref{fig:cam}, assuming that we obtain $t$ concepts from $n$ features, and the structural information of them. In this part, the quantitative knowledge from underlying data need to be mind and attached to the generated concepts. 
Logistic regression (LR) is a suitable choice. First, it is the most widely used interpretable model, and fast for inference. Second, the parameters in LR model contains the same semantic with QAF. However, the LR model only learn from the features to the prediction target, and the strength of the concept nodes and the strength of the associated edges are not directly accessible. Therefore, we propose a field-wise learning algorithm to learn values of all nodes and edges in QAF.



The field-wise learning algorithm runs in two steps. In First step, we train a LR model to learn the strength of nodes and edges of $QAF_{G}^{org}$ from original arguments (without the newly generated concepts in last process) to concept with global semantics $c_g$. $QAF_{G}^{org}$ is denoted as a QAF rooted with $c_g$ and the children only contain original arguments. In second step, we link the newly generated concept $c_i$ and its children as a sub-tree to $QAF_{G}^{org}$, and delete the edges which link children of $c_i$ directly to $c_g$, thus we get a new $QAF_{G}^{c_i}$ by adding the structure of a new generated concept $c_i$ and remove the repeated arguments. Then we use a net which has the same structure with $QAF_{G}^{c_i}$ to learn the the unknown strength of nodes and edges of $QAF_{G}^{c_i}$. The same parts of $QAF_{G}^{c_i}$ and $QAF_{G}^{org}$ has been learned in the first step, thus the net only learn strength of edges and node related to $c_i$. Hence, the learning process is `field-wise'. We repeat the second step, until all the strength of edges and nodes related to newly generated concepts are mined. And we obtain a list of fully learned QAF $(QAF_{G}^{c_1},...,QAF_{G}^{c_t})$.

Formally, denote original arguments set as $\mathcal{A}_{org} = \{a_1,..., a_q\}$, where $a$ may be features or concepts. Specially, in the first round of concept mining, $a$ only represents features. Denote children of a newly generated concept $c_i$ as $\{a_j, a_k\}$. In first step, LR model can be described as:
\begin{equation}
S(c_g) = \varphi(\sum_{a_i\in \mathcal{A}_{org}} w_i\times a_i +b_g) 
\end{equation}
where $\varphi(z) =  \frac{1}{1+exp(-z)}$ is the logistic function, $w_i \in (w_1,...,w_k)$ is the weight and $b_g$ is bias.

To represent the knowledge learned from data in form of QAF, edges between arguments and concept with global semantics $c_g$ are represented as $E={(a_1,c_g),...,(a_q,c_g)}$, $w_i \in (w_1,...,w_k)$ is the strength of edge $(a_i,c_g)$, thus function $\omega$ in $QAF_{G}^{org}$ are instantiated as $\omega((a_i,c_g)) = w_i$, where $w_i \in (w_1,...,w_k)$ and $(a_i,c_g) \in E$ . $b$ represent the initial score of $c_g$. But in QAF,  $\beta(c_g) \in [0,1]$, thus we define $\beta(c_g) = \varphi(b_g)$.
In second step, a net model can be described as:
\begin{equation}
S(c_g) = \varphi(\sum_{a_i} w_i\times a_i +b_g + w_c(\varphi(w_j^{'} \times a_j + w_k{'} \times a_k +b_c ))) 
\end{equation}
where in sum function $\sum$, $ a_i \in \mathcal{A}_{org} \setminus \{a_j, a_k\}$, and $w_i$ is learned in last step thus we fix $w_i$ as a constant score during net model training process. $w_c$ is the weight of newly generated concept $c_i$, $w_j^{'}, w_k{'}$ are new weights of $a_j$ and $a_k$, cause their structure has changed. $b_c$ is bias of $c_i$. And all the weights and biases can be represented in $QAF_G^{c_i}$ to instantiate $\omega$ and $\beta$ function.


\subsection{Knowledge reasoning: quantitative argumentation-based method}
To ensure the consistency of the strength in QAF in the learning process and inference process, we define a Net-based reasoning method as our quantitative argumentation-based method to complete knowledge reasoning. 
In Net-based reasoning algorithm, we have a strength $s(a) \in [0,1]$.  $s(a)$ is the strength value of argument $a$. The strength values are then updated by doing the following two steps for all $a \in \mathcal{A}$ from down to top until the concept with global semantics:
 \begin{flalign*}
 \mbox{\textbf{Aggregation: }} \alpha(a) := & \sum_{(b,a)\in E}\omega(b,a)\times s(b)\\
 \mbox{\textbf{Combination: }} s(a) := & \varphi (ln(\frac{\beta(a)}{1-\beta(a)})+\alpha(a))
 \end{flalign*}
 
 An instance is denoted as $x = [x_1,...,x_n]$, corresponding to all features in dataset. It is worth noting that the $x$ is preprocessed, such that $x_i \in [0,1]$. For the sake of simplicity of presentation, we take the outputs of the first round of concept mining as an example to illustrate the reasoning process of QAF. Thus, in $QAF_{G}^{org}$, the children of concept with global semantics $c_g$ are all features, represented as $\mathcal{A}_{org} = \{f_1,..., f_n\}$, and the edges are denoted as $E={(f_1,c_g),...,(f_n,c_g)}$
 
\begin{equation}
s(c_g) = \varphi (ln(\frac{\beta(a)}{1-\beta(a)})+\sum_{(f_i,c_g)\in E}\omega((f_i,c_g)) \times x_i)
\end{equation}

where $\omega$ and $\beta$ function are instantiated by field-wise learning algorithm in previous process.

By completing the inference for all samples in the evaluation dataset using the quantitative argumentation-based method, we can use resulting metrics Area-Under-Curve (AUC) to evaluate the performance of $QAF_{G}^{org}$. When the process of concept mining can generate new concepts, the reasoning method can evaluate whether the concepts are useful for decision making and thus filter out the irrelevant concepts. The process is described as follows.
\begin{flalign*}
 \mbox{If } AUC(QAF_{G}^{c_i}) \geq AUC(QAF_{G}^{org}), \mbox{then keep } c_i.\\
 \mbox{If } AUC(QAF_{G}^{c_i}) < AUC(QAF_{G}^{org}), \mbox{then drop } c_i.
\end{flalign*}

When the evaluation is over, the kept concepts and features not grouped for concept mining enter the knowledge mining process as a new round of input. CAM performs concept mining method and quantitative argumentation-based method repeatedly to generated all the import concepts from tabular data for decision making task as shown in Fig. \ref{fig:cam}. Until the concept mining method can not mining a new concept, the output of it is $QAF_{G}$. In $QAF_{G}$, every layer of important concepts are stacked until the concept with global semantics. In this case, quantitative argumentation-based method will not perform evaluation, but only act as a reasoning machine. Then, CAM are constructed by combining $QAF_{G}$ and reasoning machine. Thus, CAM can make decisions base on human-level knowledge.


\subsection{Dialogical explanation within CAM}
CAM are capable of providing the underlying structure for generating dialogical explanations for users. A user may interact with CAM by requesting an explanation of an argument (concepts or features). 
\begin{definition}
{\rm Given a }$c_g$ {\rm of an instance}\ x{,\rm \ and its } $QAF_{G} \langle \mathcal{A}, E, \beta, \omega \rangle$ {\rm with strength} $s(c_g)$ {\rm, an} argumentation dialogue {\rm between a user and CAM consists of} explanation requests $\mathcal{Q}_{(a)}$ for $a \in \mathcal{A}$ {\rm from user, to which CAM responds with} explanation $\mathcal{X}_{(a)}$
\end{definition}

Inspired by \cite{review} , we provide a simple argumentation dialogue as follows.

Let $r_1^+, r_1^- , r_2^+ , r_2^-$ be functions giving positive primary, negative primary, positive secondary and negative secondary, for any argument $a \in \mathcal{A}$:
\begin{align} 
\nonumber&\ r_1^+(a) = \mbox{because the supporting argument $a$ is $s(a)$};\\
\nonumber&\ r_1^-(a) = \mbox{because the attacking argument $a$ is $s(a)$};\\
\nonumber&\ r_2^+(a) = \mbox{and the supporting argument $a$ is $s(a)$};\\
\nonumber&\ r_2^-(a) = \mbox{and the attacking argument $a$ is $s(a)$}; \\
\nonumber&\ r_1^+(\varnothing) =  r_1^-(\varnothing) = r_2^+(\varnothing) = r_2^-(\varnothing) = \{\}. 
\end{align}
For any $S \subseteq \mathcal{A}$, if $S = \emptyset$, let $max(S) = \emptyset$; else, let $max(S) = argmax_{a \in S}(\omega(a)\times s(a))$, $sec(S) = argmax_{a \in S\setminus max(S)}(\omega(a)\times s(a))$, where $argmax$ refers to the argument $a$, at which the value of ($\omega(a)\times s(a)$) is as large as possible. Then, an \emph{argumentation dialogue} is such that for any $a \in \mathcal{A}$:
\begin{align}
\nonumber & {\rm if }\ a = c_g\ {\rm  and }\ s(c_g) <= 0.5\ {\rm and}\ \exists b \in  Att(a): \\ \nonumber
 &\mathcal{Q}(a)= \{ {\rm Why\ is }\ a\ {\rm \ evaluated\ as}\ s(a) \}\\ \nonumber
 &\mathcal{X}(a) = r_1^-(max(Att(a))) + r_2^-(sec(Att(a)) ;\ {\rm else:}\\ \nonumber
 \nonumber & {\rm if }\ a = c_g\ {\rm  and }\ s(c_g) > 0.5\ {\rm and}\ \exists b \in  Sup(a): \\ \nonumber
 &\mathcal{Q}(a)= \{ {\rm Why\ is}\ a\ {\rm \ evaluated\ as}\ s(a) \}\\ \nonumber
 &\mathcal{X}(a) = r_1^-(max(Sup(a))) + r_2^-(sec(Sup(a)) ;\ {\rm else:}\\ \nonumber
 \nonumber & {\rm if }\ a \in \mathcal{C}\ {\rm and}\ \exists b \in  Sup(a): \\ \nonumber
 &\mathcal{Q}(a)= \{ {\rm Why\ is}\ a \ {\rm \ evaluated\ as}\ s(a) \}\\ \nonumber
 &\mathcal{X}(a) = r_1^-(max(Sup(a))) + r_2^-(sec(Sup(a)) ;\ {\rm else:} \nonumber\\ \nonumber
 & {\rm if }\ a \in \mathcal{F}: \\ \nonumber
 &\mathcal{Q}(a)= \{ {\rm Why\ is}\ a \ {\rm \ evaluated\ as}\ s(a) \}\\ \nonumber
 &\mathcal{X}(a) = \{{\rm Because\ in\ this\ case,\ a\ is\ } x_a \} ; \nonumber
\end{align}

Our intuition here is that the dialogical explanation is simpler than but consistent with CAM by giving at most two paths which contributed most to concepts with global semantics. The explanation of $c_g$ may consist of its supporter or attackers which have significant impacts on $c_g$, depending on whether $c_g$ is accepted or not. This dialogue is fairly repetitive traced down other important arguments in support of the result.

\section{Experiments}

\subsection{Introduction of dataset}
We evaluate CAM with two real-world high risk application benchmark datasets denoted as Fico and Mimic3 and two in-domain anti-fraud datasets collected from two Alibaba e-commence applications denoted as data1 and data2. These datasets are medium-size with 10-100K samples and table. \ref{tb:dataset} summarizes these datasets. And the detail about the datasets will be described in Appendix.

\label{appendixTables}
\begin{table}
\begin{minipage}{1\linewidth}
\scriptsize
\centering \caption{Summary of datasets} \label{tb:dataset}
\renewcommand{\multirowsetup}{\centering}
\scalebox{0.8}{
        \begin{tabular}{cclcccc}
        \hline 
        \multirow{1}{*}{\textbf{Source Type}}
        &\multirow{1}{*}\textbf{{Name}}
        &\multirow{1}{*}{\textbf{Domain}}
        &\multicolumn{1}{c}{\textbf{\#Samples}}
			    &\multicolumn{1}{c}{\textbf{\#Features}}
			    &\multicolumn{1}{c}{\textbf{Positive rate}}\\
       \hline
        \multirow{2}{*}{\textbf{Benchmarks}} 
        &\multirow{1}{*}{Fico}
                & \multicolumn{1}{l}{Banking}
                 & \multicolumn{1}{c}{9871}
					   & \multicolumn{1}{c}{23}
						& \multicolumn{1}{c}{52.03\%}\\

         &\multirow{1}{*}{Mimic3}
                & \multicolumn{1}{l}{Healthcare}
                 & \multicolumn{1}{c}{27348}
					   & \multicolumn{1}{c}{57}
						& \multicolumn{1}{c}{9.83\%}\\

        \hline 
        \multirow{2}{*}{\textbf{In domain Dataset}} 
        &\multirow{1}{*}{data 1}
                & \multicolumn{1}{l}{E-commence}
                 & \multicolumn{1}{c}{96452}
					   & \multicolumn{1}{c}{33}
						& \multicolumn{1}{c}{3.2\%}\\
		 &\multirow{1}{*}{data2}
                & \multicolumn{1}{l}{E-commence}
                 & \multicolumn{1}{c}{98936}
					   & \multicolumn{1}{c}{65}
						& \multicolumn{1}{c}{2.0\%}\\
					
        \hline
        \end{tabular}
        }
\end{minipage}
\end{table}
\begin{table}
\begin{minipage}{1\linewidth}
\scriptsize
\centering \caption{Experimental results for four datasets conducted by CAM and other machine learning approaches using AUC(\%) metrics.} \label{t:all}
\renewcommand{\multirowsetup}{\centering}
\scalebox{0.68}{
        \begin{tabular}{lcccccccccccc}
       \hline
         \multirow{3}{*}{}
                &\multicolumn{8}{c|}{\textbf{Interpretable models}}
                &\multicolumn{4}{c}{\textbf{Black-box models}}\\
				& \multicolumn{2}{c}{\textbf{CAM}}
						& \multicolumn{2}{c}{\textbf{LR}}
                        & \multicolumn{2}{c}{\textbf{EBM}}
					    & \multicolumn{2}{c|}{\textbf{NODE-GA2M}}
						& \multicolumn{2}{c}{\textbf{MLP}}
                         & \multicolumn{2}{c}{\textbf{XGB}}\\
& \multicolumn{1}{c}{mean} & \multicolumn{1}{c}{std}  & \multicolumn{1}{c}{mean} & \multicolumn{1}{c}{std}& \multicolumn{1}{c}{mean} & \multicolumn{1}{c}{std}& \multicolumn{1}{c}{mean} & \multicolumn{1}{c|}{std}& \multicolumn{1}{c}{mean} & \multicolumn{1}{c}{std}& \multicolumn{1}{c}{mean} & \multicolumn{1}{c}{std}\\
        \hline
 
   
Fico&	\textbf{80.20}&	0.99&	79.74&	1.14&	80.09&	0.87&	77.22&	\textbf{0.81}&	77.22&	\textbf{0.81}&	79.88&	0.78\\
Mimic3&	79.76&	\textbf{0.49}&	77.66&	0.80&	80.71&	1.18&	81.22&	0.64&	72.71&	2.16&	\textbf{81.99}&	1.21\\
data1&	93.12&	0.23&	86.04&	0.49&	94.96&	\textbf{0.12}&	92.78&	0.66&	82.61&	1.85&	\textbf{97.13}&	0.40\\
data2&	96.16&	0.24&	83.94&	4.15&	96.17&	0.20&	96.67&	\textbf{0.14}&	71.54&	12.78&	\textbf{97.14}&	0.26\\
Average& 87.31&	\textbf{0.49}&	81.85&	1.64&	87.98&	0.59&	86.97&	0.56&	76.02&	4.40&	\textbf{89.04}&	0.66\\

        \hline
        
        \end{tabular}
        }
\end{minipage}
\end{table}

\subsection{The setting of experiment}

We use 80-20 splits for training and evaluation set and we repeat the experiments with five random seeds. All the datasets are for binary classification, and we use AUC as the evaluation metrics. CAM is compare against the interpretable methods of LR, EBM and NODE-GA2M, and the black-box models of MLP and xgboost (XGB)\cite{xgb}. The compared models are selected as they are commonly used classification tools for tabular data. And among them, XGB is widely regarded as the classification model with excellent performance, and EBM and NODE-GA2M are considered as the state-of-the-art in interpretable models in recent years. In Appendix, we provide the detail about data preprocessing and the hyperparameter selections for the models.

\subsection{Analysis on Classification results}

In Table. \ref{t:all}, we present the comparative results among the proposed CAM, LR, EBM, NODE-GA2M and MLP, XGB models in terms of mean and std of AUC\footnote{Mean is the average value of the 5 experimental results indicating the average performance of the model, and std is the variance of the 5 results indicating the stability of the model.}. The best experimental results are in bold font.
The results shows that CAM achieve best mean value of AUC in Fico dataset, while XGB performs best in other three datasets. In average, CAM ranks third behind the XGB and EBM. And EBM only has a small lead over CAM. As for the std value of AUC, CAM performs best in Mimic3, EBM achieve best score in data1, and NONE-GA2M runs most stably in Fico and data2. In average, CAM outperform other models as the most 
consistent model. Overall, the results show that CAM can competitive performance among all the interpretable and black-box models, and it has high stability.

\subsection{Interpretability Analysis on Fico Dataset}

Here we interpret the CAM instantiated in terms of risk prediction for Fico dataset. Fico dataset contains 10K credit bureau reports of consumers that used for predicting their loan defaulting risk. We provide the details in Appendix.

\begin{figure}
\centering
\includegraphics[scale=0.36]{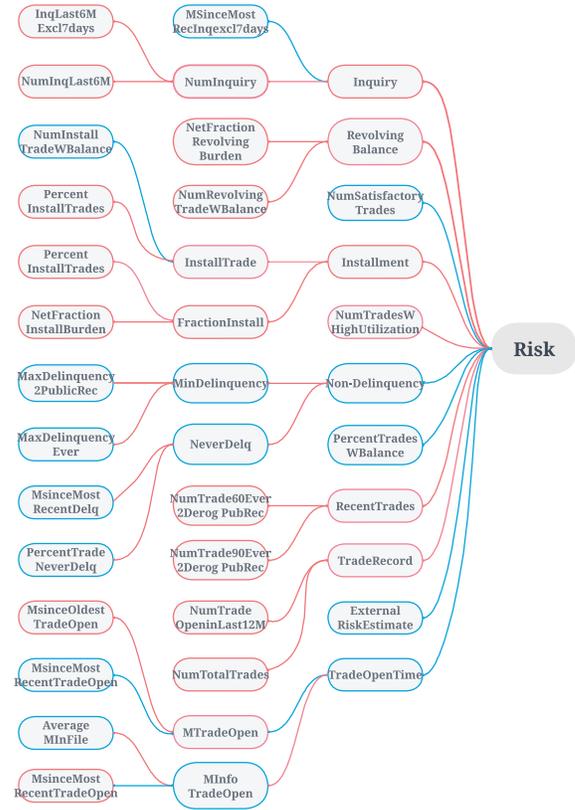}
\caption{A visualization of showing the global CAM for Fico dataset. The color of the node indicates the relation of the node to the concept risk, while the color of the edge denotes the relation of the child node to the parent node. Red and blue color indicates support and attack respectively.}
\label{fig:gobal}
\end{figure}

First, we analyze the interpretability of the global model from both semantic and argumentative perspectives. We can see that the top layer of the concept tree has 11 nodes, of which 6 concepts are generated by grouping and abstracting 18 features, and each concept can be treated as a risk factor. From the semantic perspective, Taking concept \emph{inquiry} as an example, the two child concepts \emph{MSinceMostRecInqexcl7days} and \emph{NumInquiry} describes different aspects of \emph{inquiry}, and features \emph{NumInqLast6Mexcl7days} and \emph{NumInqLast6M} describe different aspects of \emph{NumInquiry} as well. The resulted semantic tree for \emph{inquiry} is reasonable since all related features are grouped into concepts with different granularities. Additionally, We can see that in grouping, irrelevant features or concepts are not mixed in and the abstraction process makes the semantics of the nodes more general.
From an argumentative perspective, in Fig.\ref{fig:gobal}, the red and blue nodes represent the supporters and attackers to concept \emph{risk} respectively. It means that as the strength of the blue node increases, the value of the concept \emph{risk} decreases. The opposite is true for the red nodes. To be specific, we can see that when the values of feature \emph{NumInqLast6Mexcl7days} and \emph{NumInqLast6M} increase, then the value of concept \emph{NumInquiry} increase, and the value of concept \emph{Inquiry} increase, finally the value of concept \emph{Risk} increase. Combined with the semantic information in description, our observation is reasonable as it can be summarized as `when the lending institution pulled a consumer’s credit bureau report more frequently, the consumer's risk of defaulting on a loan increase'. Similarly, we can obtain other observations in line with human intuition such as `when the consumer's revolving balance increase, the consumer's risk of defaulting on a loan increase', ` when number of credit agreements on a consumer credit bureau report with on-time payments (satisfactory) increase, the consumer's risk of defaulting on a loan decrease' and etc. 

\begin{figure}
\centering
\includegraphics[scale=0.35]{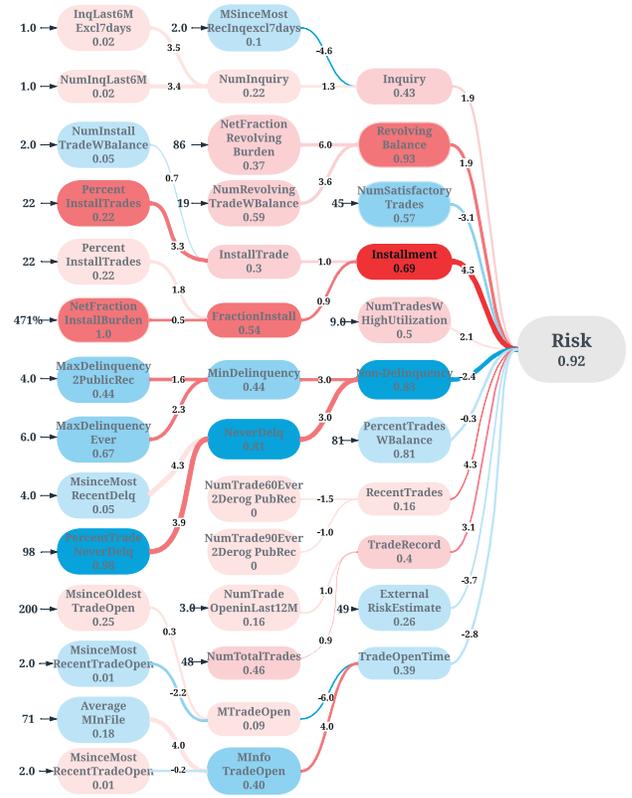}
\caption{The reasoning process of CAM on an instance. The shades of color on a node and related edge indicate the value of the product of node and edge values. A darker color indicates a larger value. And the meanings of colors remain the consistent with Fig. \ref{fig:gobal} }
\label{fig:local}
\end{figure}

Second, we start with a local example showed in Fig. \ref{fig:local}, and CAM gives the result of the reasoning and a explanation in form of dialogical tree which is showed as follows.
\begin{align}
\nonumber & \textbf{user:}\ Why\ is\ \emph{Risk}\ evaluated\ as\ 0.92? \\ 
\nonumber &  \textbf{CAM:}\ Because\ the\ supporting\ argument\ \emph{Installment}\ is \\
\nonumber & 0.69;\ and\ the\ supporting\ argument\ \emph{TradeRecord}\ is\ 0.40. \\ 
\nonumber & \textbf{user:}\ Why\ is\ \emph{Installment}\ evaluated\ as\ 0.69? \\ 
\nonumber &  \textbf{CAM:}\ Because\ the\ supporting\ argument\ \emph{FractionInstall}\ \\
\nonumber & is \ 0.54;\ and\ the\ supporting\ argument\ \emph{InstallTrade}\ is\ 0.30. \\ 
\nonumber & \textbf{user:}\ Why\ is\ \emph{FractionInstall}\ evaluated\ as\ 0.54? \\ 
\nonumber &  \textbf{CAM:}\ Because\ the\ supporting\ argument\ \\
\nonumber & \emph{FractionInstallBurden}\ is\ 1.0;\ and\ the\ \\
\nonumber & supporting\ argument\ \emph{PercentInstallTrade}\ is\ 0.22. \\ 
\nonumber & \textbf{user:}\ Why\ is\ \emph{FractionInstallBurden}\ evaluated\ as\ 1? \\ 
\nonumber &  \textbf{CAM:}\ Because\ in\ this\ case, \ \emph{FractionInstallBurden}\ is\ 471\%.
\end{align}
Users can end the conversation at any time, depending on their understanding of CAM, and he key reasoning path we can get from the above conversation is that: $x_{FractionInstallBurden}=471\% \rightarrow s(f_{FractionInstallBurden})=1 \rightarrow s(c_{FractionInstall})=0.54 \rightarrow s(c_{Installment})=0.69 \rightarrow s(c_{risk})=0.92$. With the description, we can understand the reasoning paths as `in this case, installment balance of the the consumer is 471\%  of his original loan amount, which leads the fraction of installment risk factor increases by 68.7\%. The significantly increased fraction of installment risk factor leads the installment risk factor increases by 25\% since the fraction of installment is significantly higher than other individuals. Finally, with the highest feature weights, installment risk factor contributes to the risk decision mostly'.  This explanation is very intuitive and easy for human to understand. 


\section{Conclusions and future Work}
In this work, we have proposed CAM as an interpretable model for tabular data in high-risk domains. CAM consists of a concept mining method to automatically acquire human-level knowledge as concepts in the form of QAF from both data descriptions and underlying data, and a quantitative-argumentation based method to evaluate the discovered concepts. We have also provided explanations for decisions by showing a dominated path for each reasoning process within CAM in the form of a dialogical tree.

In our experiment, we have applied CAM on four datasets in high-risk domains. The results of classification indicate that CAM can reach competitive results comparing with other state-of-the-art models. And the results of interpretation show that CAM is a global interpretable model and able to provide explanations. The knowledge inside the model is coherent with human understanding.

As a new attempt in the direction of combining knowledge mining and quantitative argumentation in interpretable research for tabular data, some aspects of this articles are still preliminary. First, the abstraction of concept names still relies on human by providing  abstracted descriptions of concepts. Second, CAM has been applied in Alibaba applications and helped users to understand the decisions, but we did not collect and evaluate users' feedback. In our future work, we are going to collect data about the feedback of the users who receive answers and explanations of their queries, and use it to our empirical study.

\subsection{References}
\nobibliography*
\bibentry{shap}.\\[.2em]
\bibentry{lime}.\\[.2em]
\bibentry{fool}.\\[.2em]
\bibentry{ebm}.\\[.2em]
\bibentry{nodegam}.\\[.2em]
\bibentry{autocross}.\\[.2em]
\bibentry{beyond}.\\[.2em]
\bibentry{deep}.\\[.2em]
\bibentry{handbookargumentation}.\\[.2em]
\bibentry{Contrastive}.\\[.2em]
\bibentry{explanation}.\\[.2em]
\bibentry{ant}.\\[.2em]
\bibentry{tweet}.\\[.2em]
\bibentry{fico}.\\[.2em]
\bibentry{tcav}.\\[.2em]
\bibentry{badgam}.\\[.2em]
\bibentry{vcec}.\\[.2em]
\bibentry{protype}.\\[.2em]
\bibentry{ace}.\\[.2em]
\bibentry{qaf}.\\[.2em]
\bibentry{baf}.\\[.2em]
\bibentry{dfqaf}.\\[.2em]
\bibentry{mlpqaf}.\\[.2em]
\bibentry{review}.\\[.2em]
\bibentry{concept}.\\[.2em]
\bibentry{xgb}.\\[.2em]
\nobibliography{aaai23}

\newpage
\section{Appendix}
\subsection{A: Details in Experiments \label{experiment}}
\paragraph{Preprocessing and acquisition of datasets}
For all datasets used in the experiment, we use 80-20 as our train-val splits, and we split all datasets with five random seeds. After that, a standardization scaler will be applied to three datasets respectively. 
For FICO dataset, we fill the missing values with the mean values and drop entirely empty rows. Link information of experiments benchmark datasets is listed in Table. \ref{link}.


For CAM training with tabular data, we target encode categorical features, apply quantile transform for all features and encode the binned features with one-hot encoder. For data description data, we first remove all special characters from the text and use pretrained multi-lingual sentence embedding models from Sentence-bert package \cite{reimers-2020-multilingual-sentence-bert}.  

\paragraph{Hyperparameters Selection}

For CAM, we use cosine distance to calculate the similarities between data descriptions and set threshold to 0.55 to generate potential concepts. For the augmentation-based field-wise filter, we freeze the models weights of the original model and fine-tune the augmentation-based model with 5 epochs. For the Logistic Regression backbone model, we use LBFGS\cite{liu89} as the optimizer and train the model with one single large batch.

For XGB, we select the hyperparameters from Node-GAM to make sure the model is fully converged. 

For LR, we use l2 regulation and the default regulation weights from the sklearn learn package. 

For EBM, we use the default parameters and set up the number of max rounds to 20K to make sure the model is fully converged.

For MLP, we uses a three layer MLP with 128 and 64 hidden units, and we apply the batch normalization\cite{Ioffe2015BatchNA} and use LeakyReLU\cite{Xu2015EmpiricalEO} as the activation function.

For NONE-GA2M, we use the second-order interaction mode and use default values for all other hyperparameters.

\begin{table}
\begin{minipage}{1\linewidth}
\scriptsize
\centering \caption{Links of Datasets.} \label{link}
\renewcommand{\multirowsetup}{\centering}
\scalebox{0.9}{
\begin{tabular}{llccccccccccc}
\hline
\textbf{Datsaset}      & \textbf{Link}\\
\textbf{FICO}          & https://community.fico.com/s/explainable-machine-learning-challenge/\\
\textbf{MIMIC3}       & https://1drv.ms/u/s!ArHmmFHCSXTIg8d6o6icAULya24iyw?e=s7TNxa/\\
\hline
\end{tabular}
}
\end{minipage}
\end{table}

\subsection{B: Details in Fico dataset \label{experiment}}
The dataset contains 23 financial features that includes trade, inquiry, delinquency, satisfactory, and utilization information. Every credit agreement between the consumer and a lending institution is represented by a separate line of information called a trade line, and is often truncated to the term `trade'. An `inquiry' is also a line of information, but captures when a lending institution has pulled a consumer’s credit bureau report in order to make a credit decision. The term `delinquency' refers to a payment received some period of time past its due date. This is typically measured in 30-day intervals, such as 60 days delinquent or 90 days delinquent. NumSatisfactoryTrades counts the number of credit agreements on a consumer credit bureau report with on-time payments (satisfactory). NumTradesWHighUtilization counts the number of credit cards on a consumer credit bureau report carrying a balance that is at 75\% of its limit or greater. The ratio of balance to limit is referred to as `utilization'. And the data description is list in Table. \ref{des}

\begin{table}
\begin{minipage}{1\linewidth}
\scriptsize
\centering \caption{Data description of Fico Datasets.} \label{des}
\renewcommand{\multirowsetup}{\centering}
\scalebox{0.7}{
\begin{tabular}{llccccccccccc}
\hline
\textbf{Feature}      & \textbf{Description}\\
\emph{ExternalRiskEstimate} 	 & Consolidated version of risk markers.\\
\emph{MSinceOldestTradeOpen}	 & Months Since Oldest Trade Open.\\
\emph{MSinceMostRecentTradeOpen}	 & Months Since Most Recent Trade Open.\\
\emph{AverageMInFile}	 & Average Months in File.\\
\emph{NumSatisfactoryTrades}	 & Number Satisfactory Trades.\\
\emph{NumTrades60Ever2DerogPubRec}	 & Number Trades 60+ Ever.\\
\emph{NumTrades90Ever2DerogPubRec}	 & Number Trades 90+ Ever.\\
\emph{PercentTradesNeverDelinquency}	 & Percent Trades Never Delinquent.\\
\emph{MSinceMostRecentDelinquency}	 & Months Since Most Recent Delinquency.\\
\multirow{2}{*}{\emph{MaxDelinquency2PublicRecLast12M}} & Max Delinquency/Public Records Last 12 Months. And value  of it \\& from  0 to 7 indicates a drop in delinquent time from 120 days+ to never. \\
\multirow{2}{*}{\emph{MaxDelinquencyEver}}	  & Max Delinquency Ever. And value of it from 0 to 7 indicates  a drop \\&  in delinquent time from 120 days+ to never.\\
\emph{NumTotalTrades}	 & Number of Total Trades (total number of credit accounts).\\
\emph{NumTradesOpeninLast12M}	 & Number of Trades Open in Last 12 Months.\\
\multirow{2}{*}{\emph{PercentInstallTrades}}	 & Percent Installment Trades. Installment is one of several equal\\& payments for something, spread over an agreed period of time.\\
\emph{MSinceMostRecentInqexcl7days}	 & Months Since Most Recent Inquiries excluding 7days.\\
\emph{NumInqLast6M}	 & Number of Inquiries Last 6 Months.\\
\emph{NumInqLast6Mexcl7days}	 & Number of Inquiries Last 6 Months excluding 7days. \\
\multirow{2}{*}{\emph{NetFractionRevolvingBurden}}	 & Net Fraction Revolving Burden. \\
& This is revolving balance divided by credit limit.\\
\multirow{2}{*}{\emph{NetFractionInstallBurden}}	 & Net Fraction Installment Burden. \\
& This is installment balance divided by original loan amount.\\
\emph{NumRevolvingTradesWBalance}	 & Number Revolving Trades with Balance.\\
\multirow{2}{*}{\emph{NumInstallTradesWBalance}}	 & Number Installment Trades with Balance. Installment is one of several \\& equal payments for something, spread over an agreed period of time.\\
\emph{NumBank2NatlTradesWHighUtilization}	 & Number of Trades with high utilization ratio.\\
\emph{PercentTradesWBalance}	 & Percent Trades with Balance.\\

\hline
\end{tabular}
}
\end{minipage}
\end{table}

\subsection{References for appendix}
\nobibliography*
\bibentry{reimers-2020-multilingual-sentence-bert}.\\[.2em]
\bibentry{liu89}.\\[.2em]
\bibentry{Ioffe2015BatchNA}.\\[.2em]
\bibentry{Xu2015EmpiricalEO}.\\[.2em]

\begin{thebibliography}{45}
\providecommand{\natexlab}[1]{#1}

\bibitem[{Baroni et~al.(2018)Baroni, Gabbay, Giacomin, and Van~der
  Torre}]{handbookargumentation}
Baroni, P.; Gabbay, D.; Giacomin, M.; and Van~der Torre, L. 2018.
\newblock Handbook of formal argumentation.

\bibitem[{Baroni et~al.(2015)Baroni, Romano, Toni, Aurisicchio, and
  Bertanza}]{qaf}
Baroni, P.; Romano, M.; Toni, F.; Aurisicchio, M.; and Bertanza, G. 2015.
\newblock Automatic evaluation of design alternatives with quantitative
  argumentation.
\newblock \emph{Argument \& Computation}, 6(1): 24--49.

\bibitem[{Borg and Bex(2021)}]{Contrastive}
Borg, A.~M.; and Bex, F. 2021.
\newblock Contrastive Explanations for Argumentation-Based Conclusions.

\bibitem[{Bouville(2008)}]{c:22}
Bouville, M. 2008.
\newblock Crime and punishment in scientific research.
\newblock arXiv:0803.4058.

\bibitem[{Caruana et~al.(2015)Caruana, Lou, Gehrke, Koch, Sturm, and
  Elhadad}]{ebm}
Caruana, R.; Lou, Y.; Gehrke, J.; Koch, P.; Sturm, M.; and Elhadad, N. 2015.
\newblock Intelligible models for healthcare: Predicting pneumonia risk and
  hospital 30-day readmission.
\newblock In \emph{Proceedings of the 21th ACM SIGKDD international conference
  on knowledge discovery and data mining}, 1721--1730.

\bibitem[{Cayrol and Lagasquie-Schiex(2005)}]{baf}
Cayrol, C.; and Lagasquie-Schiex, M.-C. 2005.
\newblock On the acceptability of arguments in bipolar argumentation
  frameworks.
\newblock In \emph{European Conference on Symbolic and Quantitative Approaches
  to Reasoning and Uncertainty}, 378--389. Springer.

\bibitem[{Chang, Caruana, and Goldenberg(2021)}]{nodegam}
Chang, C.-H.; Caruana, R.; and Goldenberg, A. 2021.
\newblock Node-gam: Neural generalized additive model for interpretable deep
  learning.
\newblock \emph{arXiv preprint arXiv:2106.01613}.

\bibitem[{Chang et~al.(2021)Chang, Tan, Lengerich, Goldenberg, and
  Caruana}]{badgam}
Chang, C.-H.; Tan, S.; Lengerich, B.; Goldenberg, A.; and Caruana, R. 2021.
\newblock How interpretable and trustworthy are gams?
\newblock In \emph{Proceedings of the 27th ACM SIGKDD Conference on Knowledge
  Discovery \& Data Mining}, 95--105.

\bibitem[{Chen et~al.(2019)Chen, Li, Tao, Barnett, Rudin, and Su}]{protype}
Chen, C.; Li, O.; Tao, D.; Barnett, A.; Rudin, C.; and Su, J.~K. 2019.
\newblock This looks like that: deep learning for interpretable image
  recognition.
\newblock \emph{Advances in neural information processing systems}, 32.

\bibitem[{Chen et~al.(2018)Chen, Lin, Rudin, Shaposhnik, Wang, and Wang}]{fico}
Chen, C.; Lin, K.; Rudin, C.; Shaposhnik, Y.; Wang, S.; and Wang, T. 2018.
\newblock An interpretable model with globally consistent explanations for
  credit risk.
\newblock \emph{arXiv preprint arXiv:1811.12615}.

\bibitem[{Chen et~al.(2022)Chen, Liao, Wan, Chen, and Wu}]{deep}
Chen, J.; Liao, K.; Wan, Y.; Chen, D.~Z.; and Wu, J. 2022.
\newblock Danets: Deep abstract networks for tabular data classification and
  regression.
\newblock In \emph{Proceedings of the AAAI Conference on Artificial
  Intelligence}, volume~36, 3930--3938.

\bibitem[{Chen and Guestrin(2016)}]{xgb}
Chen, T.; and Guestrin, C. 2016.
\newblock Xgboost: A scalable tree boosting system.
\newblock In \emph{Proceedings of the 22nd acm sigkdd international conference
  on knowledge discovery and data mining}, 785--794.

\bibitem[{Chi and Liao(2022)}]{tweet}
Chi, H.; and Liao, B. 2022.
\newblock A quantitative argumentation-based Automated eXplainable Decision
  System for fake news detection on social media.
\newblock \emph{Knowledge-Based Systems}, 242: 108378.

\bibitem[{Chi et~al.(2021)Chi, Lu, Liao, Xu, and Liu}]{ant}
Chi, H.; Lu, Y.; Liao, B.; Xu, L.; and Liu, Y. 2021.
\newblock An optimized quantitative argumentation debate model for fraud
  detection in e-commerce transactions.
\newblock \emph{IEEE Intelligent Systems}, 36(2): 52--63.

\bibitem[{Clancey(1979)}]{c:79}
Clancey, W.~J. 1979.
\newblock \emph{{Transfer of Rule-Based Expertise through a Tutorial
  Dialogue}}.
\newblock {Ph.D.} diss., Dept.\ of Computer Science, Stanford Univ., Stanford,
  Calif.

\bibitem[{Clancey(1983)}]{c:83}
Clancey, W.~J. 1983.
\newblock {Communication, Simulation, and Intelligent Agents: Implications of
  Personal Intelligent Machines for Medical Education}.
\newblock In \emph{Proceedings of the Eighth International Joint Conference on
  Artificial Intelligence {(IJCAI-83)}}, 556--560. Menlo Park, Calif: {IJCAI
  Organization}.

\bibitem[{Clancey(1984)}]{c:84}
Clancey, W.~J. 1984.
\newblock {Classification Problem Solving}.
\newblock In \emph{Proceedings of the Fourth National Conference on Artificial
  Intelligence}, 45--54. Menlo Park, Calif.: AAAI Press.

\bibitem[{Clancey(2021)}]{c:21}
Clancey, W.~J. 2021.
\newblock {The Engineering of Qualitative Models}.
\newblock Forthcoming.

\bibitem[{Cocarascu, Rago, and Toni(2019)}]{review}
Cocarascu, O.; Rago, A.; and Toni, F. 2019.
\newblock Extracting dialogical explanations for review aggregations with
  argumentative dialogical agents.
\newblock In \emph{Proceedings of the 18th International Conference on
  Autonomous Agents and MultiAgent Systems}, 1261--1269. Association for
  Computing Machinery.

\bibitem[{Engelmore and Morgan(1986)}]{em:86}
Engelmore, R.; and Morgan, A., eds. 1986.
\newblock \emph{Blackboard Systems}.
\newblock Reading, Mass.: Addison-Wesley.

\bibitem[{Eysenck(2012)}]{concept}
Eysenck, M.~W. 2012.
\newblock Fundamentals of cognition (2nd).

\bibitem[{Fang et~al.(2020)Fang, Kuang, Lin, Wu, and Yao}]{vcec}
Fang, Z.; Kuang, K.; Lin, Y.; Wu, F.; and Yao, Y.-F. 2020.
\newblock Concept-based explanation for fine-grained images and its application
  in infectious keratitis classification.
\newblock In \emph{Proceedings of the 28th ACM international conference on
  Multimedia}, 700--708.

\bibitem[{Ghorbani et~al.(2022)Ghorbani, Berenbaum, Ivgi, Dafna, and
  Zou}]{beyond}
Ghorbani, A.; Berenbaum, D.; Ivgi, M.; Dafna, Y.; and Zou, J.~Y. 2022.
\newblock Beyond Importance Scores: Interpreting Tabular ML by Visualizing
  Feature Semantics.
\newblock \emph{Information}, 13(1).

\bibitem[{Ghorbani et~al.(2019)Ghorbani, Wexler, Zou, and Kim}]{ace}
Ghorbani, A.; Wexler, J.; Zou, J.~Y.; and Kim, B. 2019.
\newblock Towards automatic concept-based explanations.
\newblock \emph{Advances in Neural Information Processing Systems}, 32.

\bibitem[{Hasling, Clancey, and Rennels(1984)}]{hcr:83}
Hasling, D.~W.; Clancey, W.~J.; and Rennels, G. 1984.
\newblock Strategic explanations for a diagnostic consultation system.
\newblock \emph{International Journal of Man-Machine Studies}, 20(1): 3--19.

\bibitem[{Hasling et~al.(1983)Hasling, Clancey, Rennels, and Test}]{hcrt:83}
Hasling, D.~W.; Clancey, W.~J.; Rennels, G.~R.; and Test, T. 1983.
\newblock {Strategic Explanations in Consultation---Duplicate}.
\newblock \emph{The International Journal of Man-Machine Studies}, 20(1):
  3--19.

\bibitem[{Ioffe and Szegedy(2015)}]{Ioffe2015BatchNA}
Ioffe, S.; and Szegedy, C. 2015.
\newblock Batch Normalization: Accelerating Deep Network Training by Reducing
  Internal Covariate Shift.
\newblock In \emph{ICML}.

\bibitem[{Kim et~al.(2018)Kim, Wattenberg, Gilmer, Cai, Wexler, Viegas
  et~al.}]{tcav}
Kim, B.; Wattenberg, M.; Gilmer, J.; Cai, C.; Wexler, J.; Viegas, F.; et~al.
  2018.
\newblock Interpretability beyond feature attribution: Quantitative testing
  with concept activation vectors (tcav).
\newblock In \emph{International conference on machine learning}, 2668--2677.
  PMLR.

\bibitem[{Kori, Glocker, and Toni(2022)}]{kori2022glance}
Kori, A.; Glocker, B.; and Toni, F. 2022.
\newblock GLANCE: Global to Local Architecture-Neutral Concept-based
  Explanations.
\newblock \emph{arXiv preprint arXiv:2207.01917}.

\bibitem[{Lakkaraju et~al.(2019)Lakkaraju, Kamar, Caruana, and
  Leskovec}]{DBLP:conf/aies/LakkarajuKCL19}
Lakkaraju, H.; Kamar, E.; Caruana, R.; and Leskovec, J. 2019.
\newblock Faithful and Customizable Explanations of Black Box Models.
\newblock In \emph{AIES}, 131--138.

\bibitem[{{LIU} and {NOCEDAL}(1989)}]{liu89}
{LIU}, D.~C.; and {NOCEDAL}, J. 1989.
\newblock On the limited memory {B}{F}{G}{S} method for large scale
  optimization.
\newblock \emph{Math. Programming}, 45(3, (Ser. B)): 503--528.

\bibitem[{Lundberg and Lee(2017)}]{shap}
Lundberg, S.~M.; and Lee, S.-I. 2017.
\newblock A unified approach to interpreting model predictions.
\newblock \emph{Advances in neural information processing systems}, 30.

\bibitem[{Luo et~al.(2019)Luo, Wang, Zhou, Yao, Tu, Chen, Dai, and
  Yang}]{autocross}
Luo, Y.; Wang, M.; Zhou, H.; Yao, Q.; Tu, W.-W.; Chen, Y.; Dai, W.; and Yang,
  Q. 2019.
\newblock Autocross: Automatic feature crossing for tabular data in real-world
  applications.
\newblock In \emph{Proceedings of the 25th ACM SIGKDD International Conference
  on Knowledge Discovery \& Data Mining}, 1936--1945.

\bibitem[{{NASA}(2015)}]{c:23}
{NASA}. 2015.
\newblock Pluto: The 'Other' Red Planet.
\newblock \url{https://www.nasa.gov/nh/pluto-the-other-red-planet}.
\newblock Accessed: 2018-12-06.

\bibitem[{Potyka(2021)}]{mlpqaf}
Potyka, N. 2021.
\newblock Interpreting neural networks as quantitative argumentation
  frameworks.
\newblock In \emph{Proceedings of the AAAI Conference on Artificial
  Intelligence}, volume~35, 6463--6470.

\bibitem[{Prakken and Ratsma(2022)}]{explanation}
Prakken, H.; and Ratsma, R. 2022.
\newblock A Top-level Model of Case-based Argumentation for Explanation:
  Formalisation and Experiments.
\newblock 159 – 194.

\bibitem[{Rago and Toni(2017)}]{polling}
Rago, A.; and Toni, F. 2017.
\newblock Quantitative argumentation debates with votes for opinion polling.
\newblock In \emph{International Conference on Principles and Practice of
  Multi-Agent Systems}, 369--385. Springer.

\bibitem[{Rago et~al.(2016)Rago, Toni, Aurisicchio, and Baroni}]{dfqaf}
Rago, A.; Toni, F.; Aurisicchio, M.; and Baroni, P. 2016.
\newblock Discontinuity-free decision support with quantitative argumentation
  debates.
\newblock In \emph{Fifteenth International Conference on the Principles of
  Knowledge Representation and Reasoning}.

\bibitem[{Reimers and Gurevych(2020)}]{reimers-2020-multilingual-sentence-bert}
Reimers, N.; and Gurevych, I. 2020.
\newblock Making Monolingual Sentence Embeddings Multilingual using Knowledge
  Distillation.
\newblock In \emph{Proceedings of the 2020 Conference on Empirical Methods in
  Natural Language Processing}. Association for Computational Linguistics.

\bibitem[{Ribeiro, Singh, and Guestrin(2016)}]{lime}
Ribeiro, M.~T.; Singh, S.; and Guestrin, C. 2016.
\newblock " Why should i trust you?" Explaining the predictions of any
  classifier.
\newblock In \emph{Proceedings of the 22nd ACM SIGKDD international conference
  on knowledge discovery and data mining}, 1135--1144.

\bibitem[{Rice(1986)}]{r:86}
Rice, J. 1986.
\newblock {Poligon: A System for Parallel Problem Solving}.
\newblock Technical Report KSL-86-19, Dept.\ of Computer Science, Stanford
  Univ.

\bibitem[{Robinson(1980{\natexlab{a}})}]{r:80}
Robinson, A.~L. 1980{\natexlab{a}}.
\newblock New Ways to Make Microcircuits Smaller.
\newblock \emph{Science}, 208(4447): 1019--1022.

\bibitem[{Robinson(1980{\natexlab{b}})}]{r:80x}
Robinson, A.~L. 1980{\natexlab{b}}.
\newblock {New Ways to Make Microcircuits Smaller---Duplicate Entry}.
\newblock \emph{Science}, 208: 1019--1026.

\bibitem[{Slack et~al.(2020)Slack, Hilgard, Jia, Singh, and Lakkaraju}]{fool}
Slack, D.; Hilgard, S.; Jia, E.; Singh, S.; and Lakkaraju, H. 2020.
\newblock Fooling lime and shap: Adversarial attacks on post hoc explanation
  methods.
\newblock In \emph{Proceedings of the AAAI/ACM Conference on AI, Ethics, and
  Society}, 180--186.

\bibitem[{Xu et~al.(2015)Xu, Wang, Chen, and Li}]{Xu2015EmpiricalEO}
Xu, B.; Wang, N.; Chen, T.; and Li, M. 2015.
\newblock Empirical Evaluation of Rectified Activations in Convolutional
  Network.
\newblock \emph{ArXiv}, abs/1505.00853.

\end{thebibliography}
\nobibliography{aaai23}

\end{document}